\documentclass[11pt]{article}
\usepackage[preprint]{latex/acl}
\usepackage{amsmath}
\usepackage{times}
\usepackage{latexsym}
\usepackage{graphicx}
\usepackage{makecell}
\usepackage{comment}
\usepackage{algorithm}
\usepackage{algorithmic}
\usepackage[T1]{fontenc}
\usepackage{algorithm}
\usepackage{algorithmic}
\usepackage[utf8]{inputenc}
\usepackage{multirow}
\usepackage{enumitem}

\usepackage{listings}
\usepackage{xcolor}

\usepackage{microtype}
\usepackage{hyperref}
\usepackage{inconsolata}
\usepackage{amssymb}
\usepackage{amsmath}
\usepackage{booktabs}
\usepackage{lipsum}
\usepackage{colortbl}

\usepackage{pifont} 
\newcommand{\cmark}{\ding{51}} 
\newcommand{\xmark}{\ding{55}} 
\newcommand{\ie}{\textit{i}.\textit{e}.}
\newcommand{\eg}{\textit{e}.\textit{g}.}
\usepackage{xspace}
\makeatletter
\DeclareRobustCommand\onedot{\futurelet\@let@token\@onedot}
\def\@onedot{\ifx\@let@token.\else.\null\fi\xspace}
\def\eg{\emph{e.g}\onedot} 
\def\ie{\emph{i.e}\onedot} 
\def\cf{\emph{c.f}\onedot}

\title{Token Preference Optimization with Self-Calibrated Visual-Anchored Rewards for Hallucination Mitigation}

\author{
  \textbf{Jihao Gu}\thanks{Equal contribution.}\thanks{Work done during an internship at Alibaba Group.}\textsuperscript{1}, 
  \textbf{Yingyao Wang}\footnotemark[1]\textsuperscript{1}, 
  \textbf{Meng Cao}\textsuperscript{2}, 
  \textbf{Pi Bu}\textsuperscript{1}, \\ 
  \textbf{Jun Song}\thanks{Corresponding Author.}\textsuperscript{1}, 
  \textbf{Yancheng He}\textsuperscript{1}, 
  \textbf{Shilong Li}\textsuperscript{1}, 
  \textbf{Bo Zheng}\textsuperscript{1} \\
  \textsuperscript{1}Taobao \& Tmall Group of Alibaba \\
  \textsuperscript{2}Mohamed bin Zayed University of Artificial Intelligence \\
  {\{gujihao.gjh, wangyingyao.wyy, jsong.sj\}}@taobao.com
}

\begin{document}
\maketitle
\begin{abstract}
Direct Preference Optimization (DPO) has been demonstrated to be highly effective in mitigating hallucinations in Large Vision Language Models (LVLMs) by aligning their outputs more closely with human preferences. Despite the recent progress, existing methods suffer from two drawbacks: 1) Lack of scalable token-level rewards; and 2) Neglect of visual-anchored tokens. To this end, we propose a novel Token Preference Optimization model with self-calibrated rewards (dubbed as TPO), which adaptively attends to visual-correlated tokens without fine-grained annotations. Specifically, we introduce a token-level \emph{visual-anchored} \emph{reward} as the difference of the logistic distributions of generated tokens conditioned on the raw image and the corrupted one. In addition, to highlight the informative visual-anchored tokens, a visual-aware training objective is proposed to enhance more accurate token-level optimization. Extensive experimental results have manifested the state-of-the-art performance of the proposed TPO. For example, by building on top of LLaVA and Qwen, our TPO boosts the performance absolute improvement for hallucination benchmarks. Our code is available at https://github.com/alibaba/TPO.
\end{abstract}

\section{Introduction}


Recently, Large Vision Language Models (LVLMs) have showcased their remarkable capabilities in handling multimodal information, excelling in tasks such as image captioning, visual question-answering, and complex visual reasoning \cite{team2023gemini,bai2023qwen,hurst2024gpt,yang2023dawn}. Specifically, by integrating pre-trained language models with meticulously designed visual encoders, LVLMs are capable of effectively capturing the semantic correlations between visual and textual data. This integration supports more accurate and contextually relevant tasks of visual understanding and generation.

\begin{figure}[!t]
\begin{center}
\includegraphics[scale=0.45]{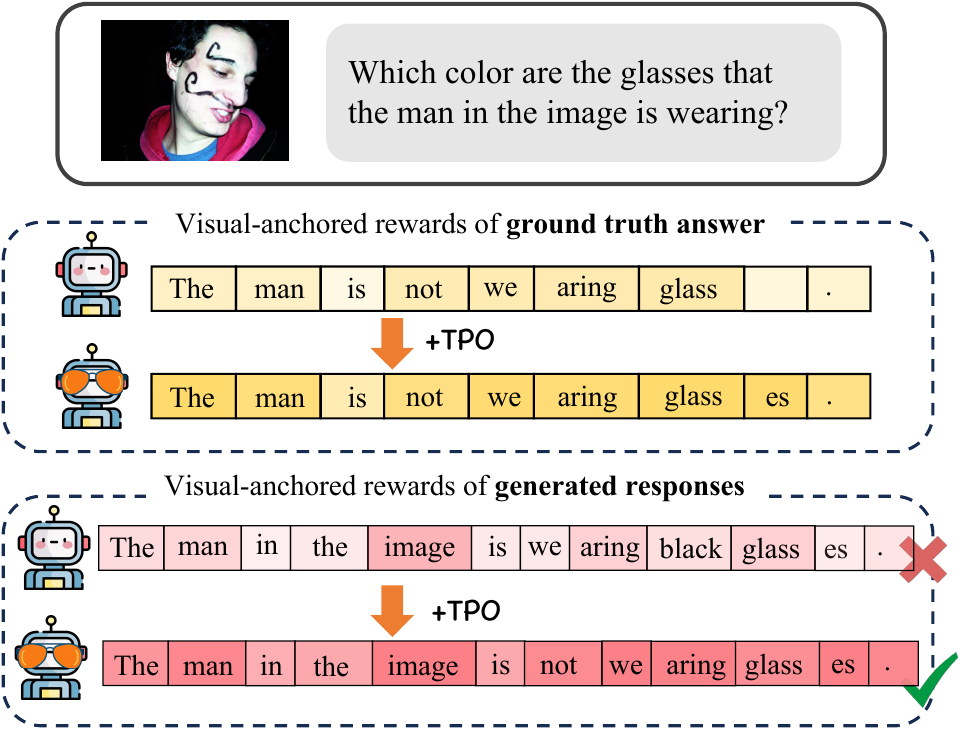}
\caption{ An example of visual Q\&A. The upper box contains the ground truth answer, while the lower box shows the LVLM responses before and after training with our method. In each box, we visualize the rewards for each token which can reflect the degree of visual anchoring, with the top representing scores before training and the bottom after. Scoring is detailed in Equation \ref{(5)}, and we've applied sigmoid normalization in this score.}
\label{fig:teaser}
\vspace{-0.5cm}
\end{center}
\end{figure}

Despite the advancements, the issue of ``hallucination'', where the generated responses are not grounded in the input visual contexts, greatly impedes the reliability and practical deployment of LVLMs \cite{liu2024survey,bai2024hallucination}. To alleviate this, various methods have been proposed from the perspectives of data quality \cite{liu2023mitigating,zhai2023halle} and inference-time strategies \cite{yin2023woodpecker,zhou2023analyzing,huang2024opera}. Recently, direct preference optimization (DPO) \cite{rafailov2024direct} is introduced to align outputs with human preferences, therefore reducing the risk of generating hallucinatory or nonsensical responses.

Existing DPO-like methods, however, still suffer from two drawbacks: 1) \emph{Lack of scalable token-level rewards}. The fine-grained token-level rewards enable precise adjustments to individual parts of generated responses. Existing methods, however, either provide global sentence rewards or rely on manual efforts for fine-grained segment-level annotations \cite{yu2024rlhf}. Therefore, designing a scalable token-level reward generation strategy has become a clearly defined necessity (\cf Table \ref{table.3}); 2) \emph{Neglect of visual-anchored tokens:} By ``visual-anchored tokens'', we refer to response tokens that are essential and highly correlated with the input visual embeddings. RLHF-V assigns all the hallucinated segments with a fixed reward value. Recent studies \cite{guan2024hallusionbench} attribute the hallucination issue to an inherent imbalance between the visual and textual modalities. Specifically, due to the large-scale pre-trained textual corpus, LVLMs tend to prioritize language-based information even at the costs of overriding the provided visual content. Therefore, we argue that not all the tokens are equal, \ie, visual-anchored tokens (\eg, \texttt{glass} in Figure \ref{fig:teaser}) are more prone to hallucination and deserve great emphasis. As shown in Table \ref{table.3}, the concurrent pre-print V-DPO \cite{xie2024v} also focuses on visual-anchored tokens; however, it requires the additional construction of a synthetic dataset, whereas our method eliminates the need for any extra annotations.



To alleviate these aforementioned problems, we propose a novel \textbf{T}oken \textbf{P}reference \textbf{O}ptimization with self-calibrated rewards (dubbed as \textbf{TPO}), which rectifies the fine-grained token-level hallucinations and attends to visual-anchored tokens without the need of fine-grained annotations. Specifically, to mine the visual-anchored tokens, we compute the differences between the logits distributions of generated tokens conditioned on the raw image and the corrupted one. We regard this distribution difference as token-wise rewards. In Figure \ref{fig:teaser}, we apply this visual-anchored score mining strategy on both golden truth and the generated responses. As shown, this strategy effectively helps highlight visual-anchored tokens. Then, we propose a token preference optimization loss by integrating the self-calibrated rewards into the vanilla DPO. In particular, we multiply the like-hood distribution with token-wise rewards to generate our desired visual-correlated ones. 

Overall, the main contributions of this work are:
\begin{itemize}[topsep=0pt, partopsep=0pt, leftmargin=13pt, parsep=0pt, itemsep=3pt]
    \item We propose TPO for hallucination mitigation in LVLMs, which implements token-level distribution rectification without the reliance of fine-grained manual annotations.

    \item We mine visual-anchored tokens by comparing the response distributions conditioned on the raw image and the corrupted one.

    \item Extensive experiments on the popular hallucination benchmarks demonstrate the state-of-the-art performance of the proposed TPO.
\end{itemize}

\begin{table}[t]
    \centering
    \small
    \begin{tabular}{ l | c c c }
        \toprule
        \textbf{Methods} & \makecell[c]{\bf Visual-\\ \bf Anchored} & \makecell[c]{\bf Token-\\ \bf level} & \makecell[c]{\bf Non Fine-grained\\ \bf Annotations}  \\
        \midrule
        DPO  & \xmark & \xmark &  \cmark\\
        POVID  & \xmark & \xmark &   \cmark\\
        CSR & \cmark & \xmark &  \cmark \\
       MDPO & \cmark & \xmark &  \cmark \\
       V-DPO & \cmark & \cmark & \xmark \\
        RLHF-V & \xmark & \cmark &  \xmark \\
         \rowcolor[gray]{0.9}\textbf{TPO (Ours)} & \cmark & \cmark & \cmark \\
        \bottomrule
    \end{tabular}
\caption{Comparisons with hallucination mitigation methods from the perspective of whether attending to vision-anchored tokens, whether generating token-level rewards and whether requiring fine-grained annotations. The compared methods include DPO \cite{rafailov2024direct}, POVID \cite{zhou2024aligning}, CSR \cite{zhou2024calibrated}, MDPO \cite{wang2024mdpo}, V-DPO \cite{xie2024v}, RLHF-V \cite{yu2024rlhf}.}
    \label{table.3}
    \vspace{-0.5cm}
\end{table}

\section{Related Works}
\subsection{LVLMs' Hallucination}
Leveraging the rich knowledge in large language models and the vision understanding capabilities of vision encoders, LVLMs have shown exceptional performance in image understanding and generation tasks \cite{li2023blip,zhu2023minigpt}. However, imbalances in parameters and data scale during pre-training can lead to LVLMs being overly influenced by biases in the language model, resulting in inadequate attention to visual information and potential hallucination issues \cite{zhou2023analyzing,zhang2024eventhallusion}. Consequently, addressing the issue of hallucinations in LVLMs has become one of the key research focuses in this field.

Previous studies have mitigated hallucinations by enhancing training data quality, refining decoding strategies, and post-processing generated responses \cite{huang2024opera,leng2024mitigating,yu2024hallucidoctor,han2024skip,chen2024halc,zhou2023analyzing,yin2023woodpecker,lee2023volcano,shao2024visual,jiang2024hallucination,yue2024less,xiao2025detecting,sarkar2024mitigating,zhao2023beyond}. 
While these methods can lead to more accurate responses, they do not fundamentally resolve the issue of inadequate visual information association in LVLMs.

\begin{figure*}[t]
\begin{center}
\includegraphics[width=1\textwidth]{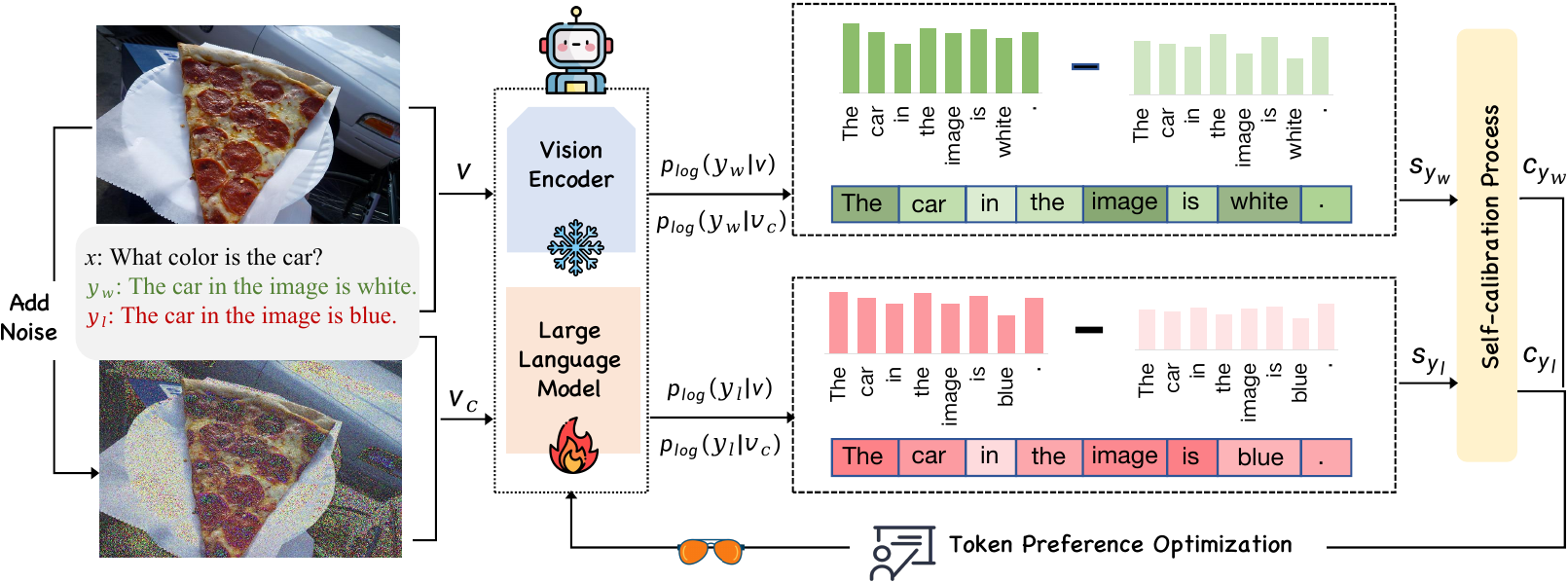}
\caption{Outline of our TPO pipline. The process is divided into three parts for each data at every training step. First, 1) add noise to the image, then, 2) calculate Self-Calibrated Visual-Anchored Rewards, and finally 3) perform Token Preference Optimization. At the end of each training step, we calibrate the model and calculate new Visual-Anchored Rewards for the next step.}
\label{fig.2}
\end{center}
\vspace{-0.5cm}
\end{figure*}

\subsection{Preference Learning Methods}
More recently, reinforcement learning from human feedback (RLHF) \cite{sun2023aligning} is gradually becoming a prevalent approach to mitigate the hallucination.
As a more direct and effective method,  DPO \cite{rafailov2024direct} and its variants are more widely utilized for preference alignment.

Several studies based on DPO focus on developing more robustly constructed preference data. For example, the POVID \cite{zhou2024aligning} method constructs negative samples for preferred data by adding noise to the image and providing hallucinated patterns to guide the model to generate hallucinated responses. 
The MDPO incorporate optimization for image preference, training with the images before and after alteration as positive and negative samples.
Apart from these works, RLAIF \cite{yu2024rlaif} and CSR \cite{zhou2024calibrated} methods, which are built upon on-policy DPO strategy, construct preference pairs by iteratively performing self-rewarding to select preference pairs. R1-Onevision \cite{yang2025r1onevision} enhances the visual reasoning capabilities by employing Group Relative Policy Optimization (GRPO). 
However, assigning response-level rewards for each generated sequence is insufficient for effectively aligning with genuinely hallucination-prone contents.

Other studies, RLHF-V \cite{yu2024rlhf} and  V-DPO \cite{xie2024v}, investigated this issue and achieved more fine-grained alignment of preference data. Nevertheless, this approach depends on resource-intensive annotations or data constructions and applies a fixed reward to all hallucinated segments, thus failing to account for the differing levels of relevance these segments may have to visual information. It is worth mentioning that CSR also considered this problem and introduced CLIP \cite{radford2021learning} to calculate the relevance score between generated text and vision information as an additional reward, and TLDR \cite{fu2024tldr} score each token by training a scoring model. However, these methods requires the introduction of an additional model, which reduces the training efficiency. 

In this paper, we propose a token-level preference optimization method with self-calibrated visual-anchored rewards (TPO), aimed at addressing the aforementioned challenges. TPO facilitates finer-grained alignment in LVLMs, enhancing accuracy in visual information correlation and reducing hallucinations during response generation.


\section{Methodology}

The schematic illustration of the proposed TPO is demonstrated in Figure \ref{fig.2}. In Sec. \ref{sec:3.1}, we present the preliminaries including the definition and off-policy optimization of DPO. Then we detail the visual-anchored rewards and token preference optimization loss in Sec. \ref{sec331} and Sec. \ref{sec:3.3}, respectively.

\subsection{Preliminaries} \label{sec:3.1}
DPO \cite{rafailov2024direct} is designed to directly maximize the reward margin between positive and negative responses to align human preferences. Given a textual input $x$, a visual input $v$, a negative response $y_l$, and a preferred positive response $y_w$, the reward function $r(x,v,y_l/y_w)$ is defined as follows.
\begin{equation}
r(x,v,y) = \beta \log \frac{\pi_\theta(y | x, v)}{\pi_{\text{ref}}(y | x, v)},  \label{(4-1)}
\end{equation}
\noindent where $\pi_{\text{ref}}{(y| x,v)}$ and $\pi_\theta(y| x,v)$ respectively represent  the reference model and current policy model. On this basis, the formulation of a maximum likelihood objective is defined as: 
\begin{equation}
\begin{aligned}
&\mathcal{L}_{DPO}(\pi_\theta; \pi_{\text{ref}}) 
= -\mathbb{E}_{(x, v, y_w, y_l) \sim D} 
 \Big[ 
    \log \sigma  \\
    &\Big( \beta \log \frac{\pi_\theta(y_w | x, v)}{\pi_{\text{ref}}(y_w | x, v)} 
    - \beta \log \frac{\pi_\theta(y_l | x, v)}{\pi_{\text{ref}}(y_l | x, v)}\Big) \Big],
\end{aligned}
\label{(4)}
\end{equation}
\noindent where $\sigma(\cdot)$ denotes the sigmoid function.
\subsection{Visual-Anchored Rewards}
\label{sec331}
Different to the equal confidence for each token in DPO, we propose a visual-anchored by measuring the token-wise visual reliance. Specifically, we firstly add noise into the embedding of the input image $v$ in a total $k$ steps to obtain the corrupted image $v_c$: 
\begin{equation}
v_c(k) = \sqrt{\bar{\xi}_k} \cdot v + \sqrt{1 - \bar{\xi}_k} \cdot \epsilon,
\label{(4)}
\end{equation}
\noindent where $\xi$ is a predefined noise parameter derived from a list with 1,000 equally spaced elements\footnote{More details can be found in Appendix \ref{A}, and experimental analysis can be found in Appendix \ref{F}}. $\bar{\xi}_k$ is a cumulant, \ie, $\bar{\xi}_k = \prod_{i=0}^{k} \xi_i$.

Subsequently, the difference of generated token distribution is computed as follow:
\begin{equation}
s_{y_i}= p_{log}(y_i | x,v,y_{<i}) 
-p_{log}(y_i | x,v_{c},y_{<i}), \label{(5)}
\end{equation}
\noindent where $s_{y_i}$ denotes the distribution difference of the token $y_i$ of the response $y$.  $p_{log}$ refers to the raw logits output of the model, before applying \texttt{softmax} normalization. 

The sole variation in the input when computing $s_{y_i}$ of the token lies in whether noise is introduced into the image. A higher contrastive score, reflecting a greater divergence in logits, indicates that the current token is more susceptible to image perturbations. This increased sensitivity suggests a stronger dependence on and relevance to visual information, thereby marking it as a visual-anchored token. One example case is demonstrated in Figure~\ref{fig:teaser}, which demonstrates that $s_{y_i}$ reflects the visual relevance of each token $y_i$.

Then, a self-calibration process is proposed to generate the final visual-anchored rewards $c_{y_i}$. 
\begin{equation}
c_{y_i}=\begin{cases}a+\sigma(s_{y_i}) & \text{if }y_i \in y_w\\
a+1-\sigma(s_{y_i}) & \text{if }y_i \in y_l
\end{cases} \label{(6)}
\end{equation}
\noindent where $a$ is a margin value. We set $a=0.5$ in Equation~\eqref{(6)}, so that when $s=0$, $c=1$, the rewards will not take effect. This process aims to ensure that positive samples receive higher rewards than negative samples while optimizing the visual relevance of visual-anchored tokens in all responses. 

\subsection{Token Preference Optimization} \label{sec:3.3}
After obtaining the reward $c_{y_i}$ to $y_i$,  the output cumulative distribution can be calculated:
\vspace{-0.12cm}
\begin{equation}
\pi^v(y | x,v)=\prod_{y_i \in \mathcal{Y}}  c_{y_i} \label{(7)}
\end{equation}
\vspace{-0.12cm}
Especially, when $c_{y_i}=1$, the probability of $y_i$ will not be accumulated. By multiplying the probability distribution with the visual-anchored rewards, we obtain a novel KL-constrained reward maximization objective:
\begin{equation}
\begin{aligned}
&\max_\pi E_{(x,v,y)} \Big[ \, r'(x,v, y) 
  - \beta D_{KL} \Big( \pi_\theta(y | x,v)   \\
  &  \cdot \pi_\theta^v(y | x,v), \pi_{\text{ref}}(y | x,v) \cdot \pi_{\text{ref}}^v(y | x,v) \Big) \Big],
\end{aligned}
\label{(8)}
\end{equation}
\noindent where $D_{KL}(\cdot, \cdot)$ denotes the KL divergency computation. $\pi_\theta^v(y | x,v)$ and $\pi_{\text{ref}}^v(y | x,v)$ are calculated using the policy model and the reference model, respectively. 
Thus, the optimal solution formula for the maximization objective of the KL-constrained reward is as follows:
\begin{equation}
\begin{aligned}
\pi_\theta(y | x,v) \cdot \pi_\theta^v(y | x,v)= \frac{1}{Z(x,v)} \pi_{\text{ref}}(y | x,v)\cdot \\
 \pi_{\text{ref}}^v(y | x,v) \exp\big(\frac{1}{\beta} r'(x, v,y)\big).
\end{aligned}
\label{(9)}
\end{equation}
The partition function of Eq \eqref{(9)} is as follows.
\begin{equation}
\begin{aligned}
    Z(x,v) = \sum_y & \pi_{\text{ref}}(y | v, x) \cdot \pi_{\text{ref}}^v(y | x,v) \\
& \cdot \exp\big(\frac{1}{\beta} r'( x,v, y)\big)
\end{aligned}
\label{(9.1)}
\end{equation}
Rearranging Eq \eqref{(9)}, we obtain the reward function:
\begin{equation}
\begin{aligned}
&r'(x,v,y) \\
&= \beta \log \frac{\pi_\theta(y | x,v)\cdot\pi_\theta^v(y | x,v)}{\pi_{\text{ref}}(y | x,v)\cdot\pi_{\text{ref}}^v(y | x,v)} + \beta Z(x,v) \\
          &= \beta \sum_{y_i \in y} \left[ \log \left( p_\theta(y_i | x, v,y_{<i}) \cdot \textcolor{red}{c_{y_i}^{\theta}} \right) \right. \\
          &\quad \left. - \log \left( p_{\text{ref}}(y_i | x, v,y_{<i}) \cdot \textcolor{blue}{c_{y_i}^{\text{ref}}} \right) \right] + \beta Z(x,v) \\
&= \beta \sum_{y_i \in y} \Big[ \log p_\theta(y_i | x, v, y_{<i}) - \log p_{\text{ref}}(y_i | x, v, y_{<i}) \\
&\quad + \log  \frac{\textcolor{red}{c_{y_i}^{\theta}}}{\textcolor{blue}{c_{y_i}^{\text{ref}}}}  \Big] + \beta Z(x,v),
\end{aligned}
\label{(10)}
\end{equation}
where $c_{y_i}^{\theta}$ and $c_{y_i}^{\text{ref}}$ represent the token reward calculated using the policy model and the reference model, respectively. 

Compared to the original reward function in DPO (Eq~\eqref{(4-1)}), we multiply each $p(y_i | x, v,y_{<i})$ by the generated visual-anchored rewards $c_{y_i}$ at the token level. $c_{y_i}^{\theta}$ is continuously updated at each step during training as the model changes. To calculate each token in the entire reward function, we add a term $log\frac{c_{y_i}^{\theta}}{c_{y_i}^{\text{ref}}}\in(-log3,log3)$ (as we set $a=0.5$ in Equation~\eqref{(6)}), which has a reasonable upper and lower bound. For positive samples, this term is expected to increase, while for negative samples, it is expected to decrease.
Due to the different methods of calculating $c_{y_i}$ that we set in Eq~\eqref{(6)}, this will encourage the increase of $s_{y_i}$ during the training process, making the token generation focus more on visual information.

Thus, following the Bradley-Terry model, when given the positive and negative samples $\mathcal{D} = \{ x^{(k)}, v^{(k)},y_w^{(k)}, y_l^{(k)} \}_{k=1}^{N}$, we obtain our maximum likelihood objective:
\begin{equation}
\begin{aligned}
&\mathcal{L}_{TPO}(\pi_\theta; \pi_{\text{ref}}) = -\mathbb{E}_{(x, v,y_w, y_l) \sim D} \Big[ \log \sigma \\
&\quad \Big( \beta \log \frac{\pi_\theta(y_w \big| x,v) \cdot \pi_\theta^v(y_w \big| x,v)}{\pi_{\text{ref}}(y_w \big| x,v) \cdot \pi_{\text{ref}}^v(y_w \big| x,v)} - \\
&\quad \beta \log \frac{\pi_\theta(y_l \big| x,v) \cdot \pi_\theta^v(y_l \big| x,v)}{\pi_{\text{ref}}(y_l \big| x,v) \cdot \pi_{\text{ref}}^v(y_l \big| x,v)} \Big) \Big] \\
& =\mathcal{L}_{DPO}(\pi_\theta; \pi_{\text{ref}})+ \mathbb{E}_{(x, v,y_w, y_l) \sim D} \Big[ \log \sigma \\
&\quad \Big( \beta \log \frac{\pi_\theta^v(y_w \big| x,v)}{\pi_{\text{ref}}^v(y_w \big| x,v)} -\beta \log \frac{\pi_\theta^v(y_l \big| x,v)}{\pi_{\text{ref}}^v(y_l \big| x,v)} \Big) \Big] \\
\end{aligned}
\label{(11)}
\end{equation}
According to Eq \eqref{(10)}, we can deduce as follows. 
\begin{equation}
\begin{aligned}
& \mathcal{L}_{TPO}(\pi_\theta; \pi_{\text{ref}}) = -\mathbb{E}_{(x, v,y_w, y_l) \sim D} \Big[ \log \sigma \\
&\quad \Big(\beta \sum_{y_{w_i} \in y_w} \Big[ \log \left( p_\theta(y_{w_i} | x, v,y_{w_{<i}}) \right) \\
&\quad - \log p_{\text{ref}}(y_{w_i} | x, v,y_{w_{<i}}) + \log \frac{c_{y_{w_i}}^{\theta}}{c_{y_{w_i}}^{\text{ref}}} \Big] \\
& + \sum_{y_{l_i} \in y_l} \Big[  \log \left( p_\theta(y_{l_i} | x, v,y_{l_{<i}}) \right) \\
&\quad - \log  p_{\text{ref}}(y_{l_i} | x, v,y_{l_{<i}}) +\log \frac{c_{y_{l_i}}^{\theta}}{c_{y_{l_i}}^{\text{ref}}}  \Big] \Big)\Big]\\
\end{aligned}
\label{(11)}
\end{equation}
where $c_{y_{w_i}}^{\theta}$ and $c_{y_{w_i}}^{\text{ref}}$ represent the token reward calculated for $y_w$ using the policy model and the reference model, respectively. The same applies to $c_{y_{w_i}}^{\theta}$, $c_{y_{w_i}}^{\text{ref}}$ and $y_l$.

\begin{table*}[t]
    \centering
    \fontsize{9.5pt}{12.5pt}\selectfont
    \begin{tabular}{p{2.3cm}p{0.7cm}p{0.7cm}p{0.8cm}p{0.85cm}p{0.75cm}p{0.75cm}p{0.8cm}p{0.5cm}p{0.6cm}p{0.7cm}p{1.2cm}}  
        \hline
        \multirow{2}{*}{\textbf{Method}} & \multicolumn{2}{c}{\textbf{AMBER}} & \multicolumn{2}{c}{\textbf{MMHal}} & \multicolumn{3}{c}{\textbf{HallusionBench}} &\multicolumn{4}{c}{\textbf{General Benchmarks}}\\
        \cmidrule(r){2-3}  \cmidrule(r){4-5} \cmidrule(r){6-8} \cmidrule(r){9-12}
        & Acc  & F1  & Score  & Hal $\downarrow$
        & Easy& Hard& aAcc  & SEED & MMB & LLaVA & MMVet\\
        \hline
        R1-Onevision & 80.2& 	85.7& 	3.85& 	36.46& 	63.74& 	50.47& 	62.80 & 35.2	 &--	 &83.7	 &67.8 \\
        \hline
        LLaVA-1.5-7B & 71.7	&74.3	&2.01	&61.46&	42.64	&41.16	&47.21 & 66.1	 &{73.3}	 &65.6	 &31.6 \\
        + DPO & {77.5} &	{82.1}	&2.14	&58.33&37.36&	37.21&	43.84 & {66.4}& 	{73.3}& 	{69.1}& 	31.6 \\
        + CSR & 73.2	&76.1	&2.05	&60.42&	\textbf{43.08}&	41.16&	47.48 & 65.9 & 	73.0 & 	68.9 & 	31.0\\
        + POVID & 71.9	&74.7	&{2.26}	&{55.21}&	{42.86}&	41.63	&47.56 & 66.1 & 	73.2 & 	68.2 & 	{31.7}\\
        + RLHF-V & 74.8	&78.5&	2.02	&60.42&	42.20	&{43.72}&	48.27 & 66.1&	73.1&	68.0&	32.3\\
        + MDPO & --	&--	&2.39	&54.00& -- &-- & -- & -- & -- & --& --\\
        + V-DPO & --	&81.6	&2.16	&56.00& -- &-- & \textbf{51.63} & -- & -- & --& --\\
         \rowcolor[gray]{0.9}+ \textbf{TPO (Ours)} & \textbf{79.3}	&\textbf{85.0}&	\textbf{2.47}	&\textbf{51.04}&41.76&	\textbf{48.37}&	{50.22} & \textbf{66.6}	 &\textbf{73.6}	 &\textbf{70.2} &	\textbf{33.0}\\
        \hline
        LLaVA-1.5-13B & 71.3&	73.1&	2.38	&53.13&	{44.40}	&36.51&	46.94& 68.2	&{76.7}&	{73.1}	&36.1\\
        + DPO & {83.2}&	{86.9}&	2.47&	{51.04}&	\textbf{45.49}&	{43.49}	&50.22 &{68.6} &	76.6	 &72.8 &	{37.5}\\
        + RLHF-V & 79.2	&82.3		&{2.50}	&	52.08	&	43.96	&	40.00		&48.27 &68.2&	{76.7}&	\textbf{76.7}	&\textbf{38.5}\\
         \rowcolor[gray]{0.9}+ \textbf{TPO (Ours)} & \textbf{83.9}	&\textbf{88.0}	&\textbf{2.72}&	\textbf{45.83}&	{44.40}	&\textbf{46.05}&	\textbf{50.93} & \textbf{68.7} & 	\textbf{76.8} & 	72.8 & 	36.2\\
        \hline
        Qwen2-VL-7B & 86.5 & \textbf{90.0}  & 3.5  & 29.0 & 67.0 & 48.8 & 64.0& \textbf{45.0}&\textbf{79.0} & 82.4 &61.4\\
        + DPO & 86.5 & \textbf{90.0}  & 3.7  & 28.1 & 67.3 & 49.3 & 64.5& \textbf{45.0}& \textbf{79.0}& 81.9 &60.2\\
         \rowcolor[gray]{0.9}+ \textbf{TPO (Ours)} & 86.4 & 89.9 & \textbf{4.2} & \textbf{18.8} & \textbf{67.9} & \textbf{50.0} & \textbf{65.2}& \textbf{45.0}& \textbf{79.0}&\textbf{82.9} &\textbf{61.4}\\
        \hline
    \end{tabular}
    \caption{Performence of LLaVA-1.5 on hallucination and general benchmarks. Score and Hall refer to the overall GPT-4 \cite{achiam2023gpt} score and hallucination rate, respectively. Easy represents the accuracy of with original images, hard represents the accuracy with manually edited challenging images, and aAcc is the average accuracy for each question. The results for POVID \cite{zhou2024aligning} and CSR \cite{zhou2024calibrated} are based on our testing of their open-source model weights, while the results for V-DPO \cite{xie2024v}, MDPO \cite{wang2024mdpo} are taken from previous work}.
    \label{table.1}
    \vspace{-0.5cm}
\end{table*}

\section{Experiment}
\subsection{Setup}
Aligning with previous DPO-based approaches on hallucination mitigation, we mainly adopt the popular LVLM, LLaVA-1.5 \cite{liu2024improved}, as the backbone model to validate the effectiveness of our TPO. Furthermore, to evaluate the effectiveness of TPO on more advanced and powerful model, we implement TPO training based on Qwen2-VL \cite{wang2024qwen2}, and compare it with the DPO method. For the dataset, we directly utilize the preference pairs provided by RLHF-V (5K) without their fine-grained human annotations. 

\paragraph{Benchmarks} 
We primarily conduct the experiments on three hallucination benchmarks: AMBER \cite{wang2023llm}, MMHal-Bench \cite{sun2023aligning}, and HallusionBench \cite{guan2024hallusionbench}. In this section, we mainly focus on AMBER's discriminative task and report the accuracy and F1 metrics referencing \cite{yu2024rlaif}. In addition, we provide the results of its Chair metric in Appendix \ref{E}.
Moreover, we also evaluate the performance of TPO on four general benchmarks: SEED Bench \cite{li2023seed}, MMBench \cite{liu2025mmbench}, LLaVA Bench \cite{liu2024visual} and MM-Vet \cite{yu2023mm}. These benchmarks are used to evaluate the performance of the models on general tasks after hallucination alignment.

\paragraph{Baselines} We mainly compare TPO with the R1-Onevision \cite{yang2025r1onevision}, LLaVA-1.5-7B SFT model, as well as with the V-DPO \cite{xie2024v} and DPO methods trained using RLHF-V \cite{yu2024rlhf} data, along with two improved methods, CSR \cite{zhou2024calibrated} and POVID \cite{zhou2024aligning}. Moreover, to evaluate the effectiveness and robustness of TPO as the model size increases, we further evaluate the performance of TPO on the LLaVA-1.5-13B model and compared it with DPO. Additionally, to demonstrate the advantages of TPO, we reproduced the strong baseline method, RLHF-V, on LLaVA-1.5-13B and conducted a comparison.
Furthermore, we additionally employ Qwen2-VL-7B \cite{wang2024qwen2} as the baseline mode and compare our TPO with DPO.

\subsection{Main Results}

In Table \ref{table.1}, we present the main results of our TPO and baselines. In hallucination benchmarks, our method shows significant improvements over all previous methods for both the 7B and 13B models, surpassing even the GRPO-based \cite{shao2024deepseekmath} R1-Onevision model. Specifically, compared to the original LLaVA model, we achieve improvements of 20.4 $\%$ on AMBER F1, 22.8$\%$ on the MMHAL score, and 8.5$\%$ on HallusionBench aAcc at most. This validates the effectiveness of our method in helping the model mitigate hallucination issues and improve the performance of visual question answering.

Notably, on the HallusionBench evaluation metrics, "Easy" represents the accuracy of original image-based questions, which tend to rely on prior knowledge, while "Hard" represents the accuracy of questions based on manually edited images, which tend to rely on visual information. Our method leads to the most significant improvement for the original model on hard questions. This indicates that our approach enables the model to focus more on visual information rather than textual prior knowledge to provide accurate answers.

In general benchmarks, our approach remains stable against the backbone models and achieves improvement on most benchmarks. We attribute it to our method helping the model associate with more visual information when answering questions. This shows that our approach can improve hallucination issues while maintaining good performance in general evaluation tasks. 

\subsection{Results on Qwen2-VL-7B} 
As Table~\ref{table.1} shown, we report the results on the key metrics of three hallucination benchmarks. 
The results indicate that our TPO outperforms DPO on most benchmarks. On Qwen2-VL-7B, which has strong inherent capabilities, using 5K RLHF-V data for DPO alignment barely improves the performance. However, introducing TPO leads to a significant further enhancement. This demonstrates that TPO can capture and learn more subtle preferences from the data and brings higher data utilization efficiency. 
In addition, the results on the chair metric in Figure \ref{chair} further demonstrate that TPO can also significantly solve the object hallucination problem of Qwen2-VL-7B.

\begin{figure*}[t]
\begin{center}
\includegraphics[scale=0.58]{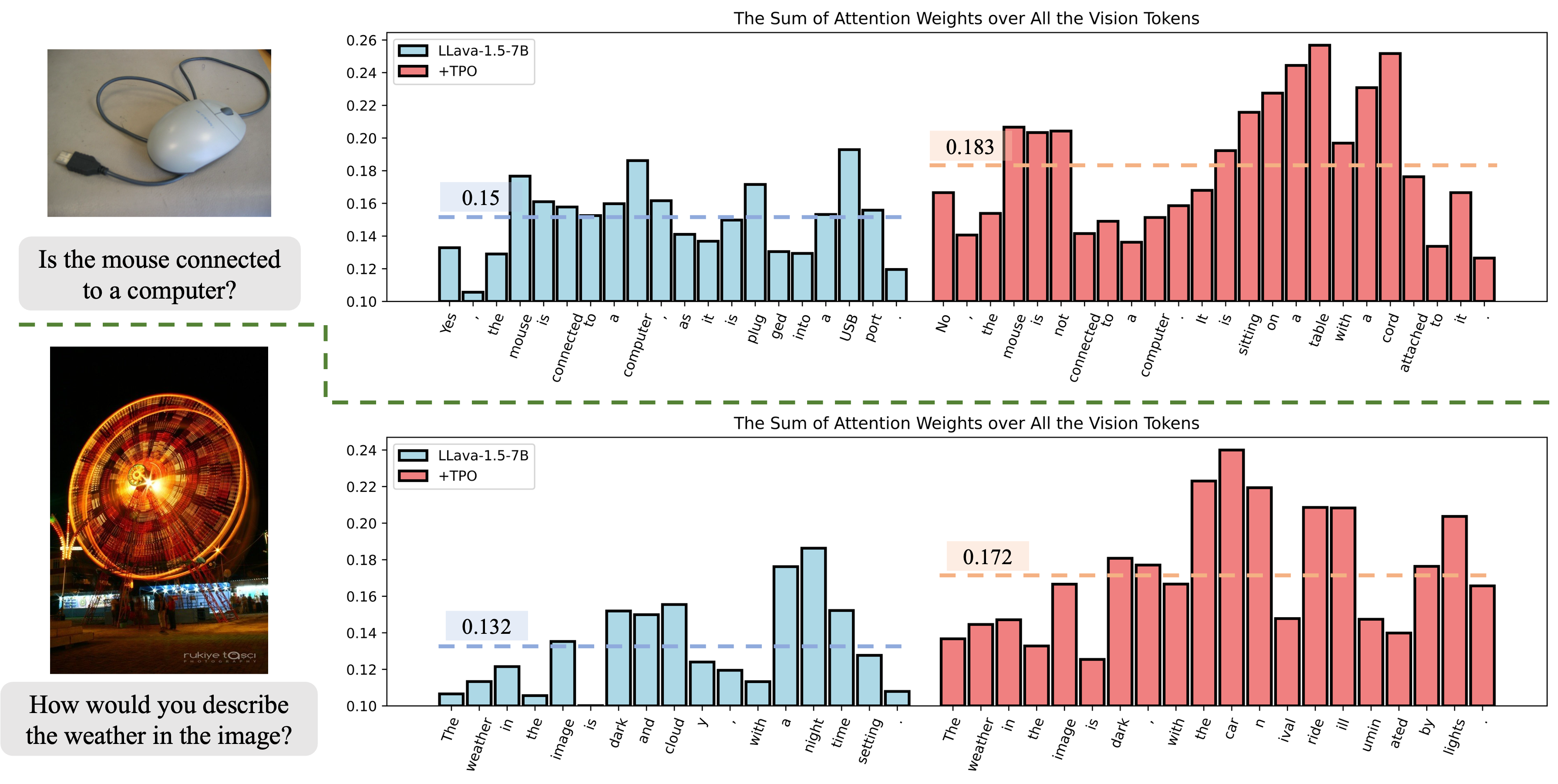}
\caption{Comparison of attention weights for LLaVA before and after TPO training. Each horizontal line represents the mean of that data. The blue section response incorrectly, with many 'visual-anchored tokens' tokens having high attention weights but resulting in hallucinated responses (\eg \texttt{USB}). The red section on the right answered correctly.}
\label{fig:attention}
\vspace{-0.5cm}
\end{center}
\end{figure*}
\begin{figure}[t]
\begin{center}
\includegraphics[scale=0.35]{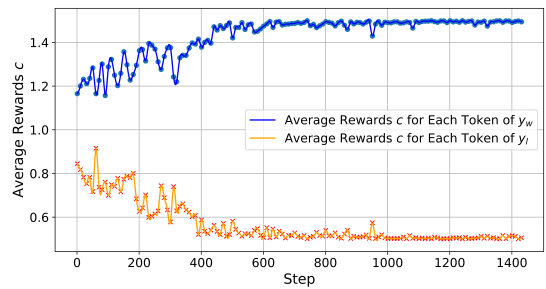}
\caption{The curve of changes in self-calibrated rewards for positive and negative samples over training steps, with a sample point taken every 10 steps.}
\label{fig:curve}
\vspace{-0.5cm}
\end{center}
\end{figure}
\subsection{Ablation Studies}
\paragraph{Visual-Anchored Rewards} Table \ref{table.2} demonstrates that TPO can enhance model performance when rewards are assigned separately to positive and negative samples, achieving results comparable to those obtained by rewarding both simultaneously. However, by providing opposite rewards to positive and negative samples, where rewards are negatively correlated with the visual relevance of positive samples and positively correlated with that of negative samples, TPO's performance significantly deteriorates. In some metrics, this approach yields even poorer results than the original LLaVA-1.5 model. This further underscores the validity of the designation of visual-anchored rewards.


\paragraph{Noise Step} We conduct an ablation study on the noise steps, as illustrated in Figure \ref{fig:step} (a) and detailed in Table \ref{table.4} of Appendix. The results indicate that optimal performance is achieved at 500 steps. This intermediate level of corruption allows the model to retain the general structure of the image while obscuring fine-grained details, thereby reducing the tendency to generate hallucinations among visual-anchored tokens.

These findings are consistent with previous work \cite{zhou2024aligning} which noted that as image noise increases, models tend to produce responses such as “there are noise spots in the image,” while the probability of hallucination first increases and then decreases. Our experimental results align with this trend and provide further empirical support.

The Figure also shows when step=0, TPO still effective and significantly better than DPO. This confusion is a code-implementation issue. In implementation as shown in Listing~\ref{lst:noise-addition}, we first convert the image into a tensor, add noise, and then convert it back into an image. This encode-decode process introduces some losses. Our method of setting the noise step to 0 serves as an ablation experiment to test the impact of this loss on our method, and it allows our experiment to more comprehensively demonstrate the advantages of TPO. The following portion of code may help you better understand our encode-decode process for adding noise. We will also open source all the code once the paper is accepted.

\paragraph{Parameter $\bf a$} We conduct experiments by varying the parameter $a$ introduced in Equation \eqref{(6)} with the results shown in Figure \ref{fig:step} (b) (detailed in Table \ref{table.5} of Appendix). By setting $a=[0,0.5,1]$, we observed consistently good performance across all configurations. This suggests that effective performance is achieved as long as the reward mechanism successfully highlights token differences and identifies visually anchored tokens. Notably, the best overall results are obtained with $a=0.5$, validating our proposed method and hypothesis. This indicates that when the visual-anchored score $s=0$, setting $c=1$, not introducing additional reward signals can yield better outcomes.

\begin{figure}[t]
\begin{center}
\includegraphics[scale=0.35]{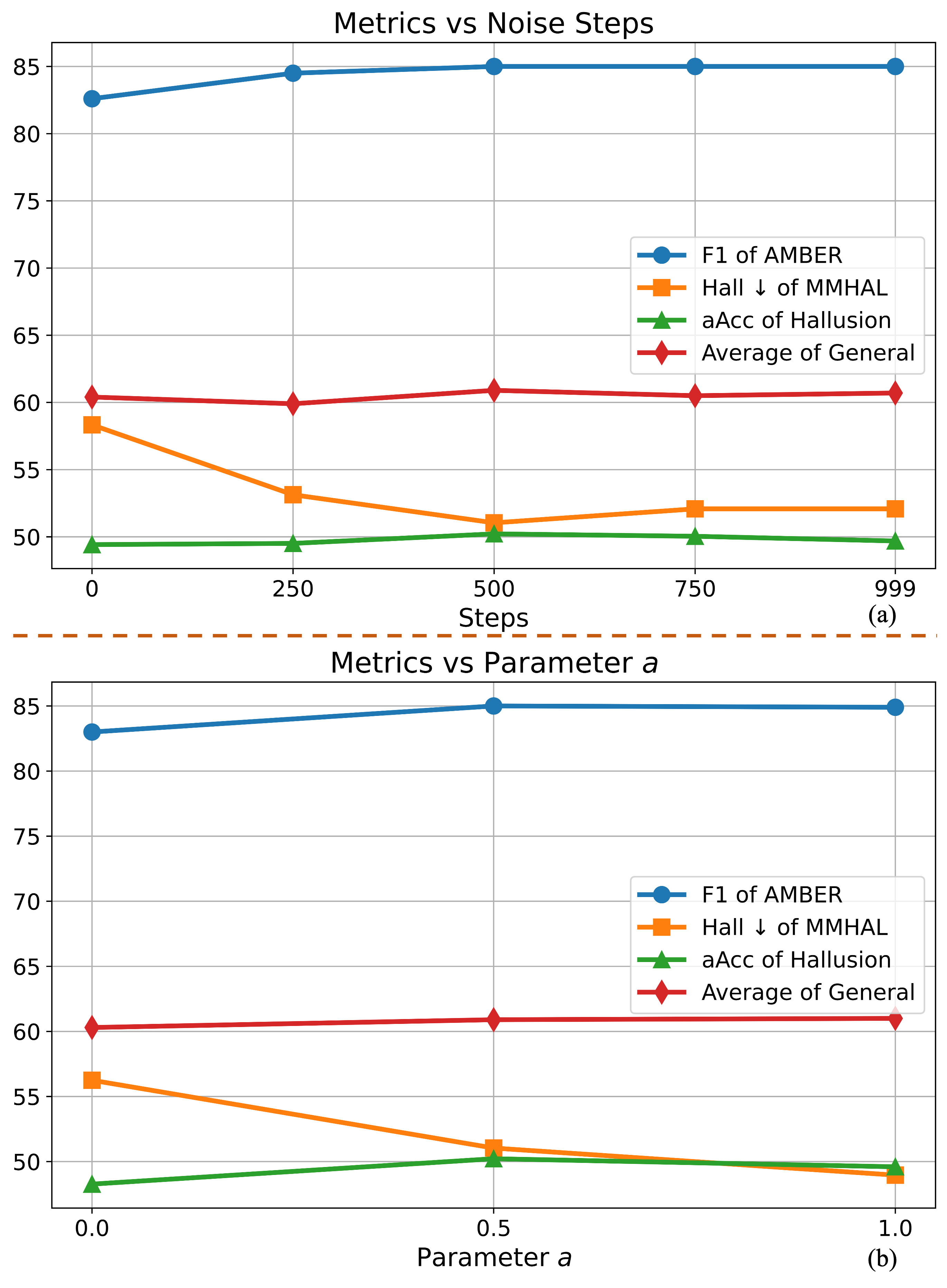}
\caption{Performance curves with the change of the noise steps-(a) and the change of parameter $a$-(b), We separately present the F1 of AMBER, the hallucination rate of MMHAL, the aACC of HallusionBench, and the average value of the general benchmarks.}
\vspace{-0.8cm}
\label{fig:step}
\end{center}
\end{figure}
\begin{table*}[t]
    \centering
    \fontsize{9.5pt}{12.5pt}\selectfont
    \begin{tabular}{p{2.3cm}p{0.7cm}p{0.7cm}p{0.8cm}p{0.85cm}p{0.75cm}p{0.75cm}p{0.8cm}p{0.5cm}p{0.6cm}p{0.7cm}p{1.2cm}}  
        \hline
        \multirow{2}{*}{\textbf{Method}} & \multicolumn{2}{c}{\textbf{AMBER}} & \multicolumn{2}{c}{\textbf{MMHal}} & \multicolumn{3}{c}{\textbf{HallusionBench}} &\multicolumn{4}{c}{\textbf{General Benchmarks}}\\
        \cmidrule(r){2-3}  \cmidrule(r){4-5} \cmidrule(r){6-8} \cmidrule(r){9-12}
        & Acc  & F1  & Score  & Hal $\downarrow$
        & Easy& Hard& aAcc  & SEED & MMB & LLaVA & MM-Vet\\
        \hline
        LLaVA-1.5-7B & 71.70	&74.3	&2.01	&61.46&	42.64	&41.16	&47.21 & 66.1	 &\underline{73.3}	 &65.6	 &31.6 \\
        \hline
        Only Win& 79.10& 	84.5& 	2.24& 	56.25& 	\textbf{44.62}& 	46.05& 	\textbf{50.40}& 66.6&	73.6&	69.8&	31.7\\
        Only Loss & 79.20&	84.8&	2.33&	53.13&	42.20	&47.91&	49.87 & 66.6&	73.5&	\textbf{70.7}&	32.0\\
        Opposite & 75.30&	80.7&	1.91&	64.58&	42.42&	45.58&	48.63&65.6&	73.1&	68.9&	32.1\\
        \hline
        \rowcolor[gray]{0.9}\textbf{TPO (Ours)} & \textbf{79.30}	&\textbf{85.0}&	\textbf{2.47}	&\textbf{51.04}&41.76&	\textbf{48.37}&	50.22 & \textbf{66.6}	 &\textbf{73.6}	 &70.2 &	\textbf{33.0}\\
        \hline
    \end{tabular}
    \caption{Ablation Studies. Performence of LLaVA-1.5 on hallucination and general benchmarks.}
    \label{table.2}
    \vspace{-0.2cm}
\end{table*}

\begin{table}[t]
   \fontsize{9pt}{12pt}\selectfont
    \centering
    \begin{tabular}{lcc}
        \hline
        \textbf{Average score} & \textbf{Noun/Adj} & \textbf{Others} \\
\hline 
        {Ground Truth} & 1.83 & 0.90 \\
        \rowcolor[gray]{0.9}{Ground Truth (TPO)}& \bf 5.72 & 4.87 \\

        {Response of LLaVA} & 1.48 & 0.83 \\
        \rowcolor[gray]{0.9}{Response of LLaVA+TPO (TPO)} & \bf 5.67 & 4.59 \\
        \hline
    \end{tabular}
    \caption{Average score from Equation~\ref{(5)} of Noun/Adj tokens and other tokens.  Here, Ground Truth  and Ground Truth  (TPO) represent the scores calculated for the ground truth answer using LLaVA-1.5-7B and LLaVA-1.5-7B+TPO. Response of LLaVA and LLaVA+TPO (TPO) correspond to the outputs before and after TPO training and the scores calculated by LLaVA-1.5-7B and LLaVA-1.5-7B+TPO, corresponding to Fiure \ref{fig:teaser}. }
    \label{tab:pos_tagging}
    \vspace{-0.5cm}
\end{table}

\subsection{Analysis}
\paragraph{Visual-Anchored Rewards}
As Figure~\ref{fig:teaser} shown, the proposed visual-anchored rewards can reflect the degree to which a token depends on visual information. To further prove this statement, we construct the analysis experiment on the MMhal dataset as shown in Table~\ref{tab:pos_tagging}. Intuitively, nouns and adjectives in responses are thought to most associate the content of an image. Therefore, we first perform part-of-speech (POS) tagging on the model responses and count the average number of noun/adjective tokens and other types of tokens. 
Specifically, in the ground-truth responses, 39.6\% of the tokens are nouns or adjectives. In the responses from LLAVA-1.5-7B, the proportion of noun and adjective tokens remains nearly constant at 39.2\%, both before and after TPO. 

Afterwards, we count the average score from Equation~\ref{(5)} of noun/adjective tokens and other types of tokens. 
The results show that noun and adjective tokens have significantly higher scores than other types, indicating higher relevance to images. After applying TPO, these scores of all the tokens increased notably. The results supports our conclusions:
1) The visual-anchored rewards reflects token-image relevance.
2) TPO enhances the alignment of generated tokens with image content.

\paragraph{Attentions}
To further validate TPO's effectiveness in enhancing visual alignment, we measure the relevance using the sum of attention weights between responses and images. On the MMHal dataset, the overall image attention weights for LLaVA-1.5-7B increased from \textbf{0.14} before TPO training to \textbf{0.17} afterward. 
Additionally, Figure \ref{fig:attention} visualizes the cases, showing a significant increase in image attention weights for response tokens, especially for visual-anchored tokens (e.g., \texttt{table}, \texttt{cord}). This highlights our method's success in improving the model's integration of visual information, thus reducing hallucinations.

\paragraph{Self-Calibration}
To illustrate that our method enables the model to progressively enhance its focus on visual information through continuous self-calibration during training, we present the evolution of scores for positive and negative samples, as calculated by Equation \eqref{(6)}, across various training steps.
With $a=0.5$, it follows that $c_{y_i} \in (0.5,1.5)$. As shown in Figure~\ref{fig:curve}, the scores for positive samples gradually approach their maximum values, while those for negative samples approach their minimum values, indicating convergence. This trend illustrates the self-calibrating effect of our method, which ultimately enhances the model's ability to focus on visual information.

\section{Conclusion}
In this study, we propose a novel pereference alignment method, TPO, to mitigation hallucinations in LVLMs. TPO incorporates a self-calibrated visual-anchored reward mechanism that automatically identifies "vision-anchored tokens" and adaptively assigns appropriate rewards to them. 
By adding noise to the visual input and capturing changes in the generation probability of each token, TPO computes a score indicating each token's relevance to visual information. 
Based on the self-calibrated visual-anchored reward, TPO can perform more efficient token-level preference alignment optimization for LVLMs. Experimental results have proved that TPO not only alleviates the hallucination problem but also strengthens the model's attention to visual input when generating responses.

\section{Limitation}
Although our method has achieved outstanding performance in addressing the hallucination problem, the self-calibrated visual-anchored rewards approach we used in this paper can be extended to even broader areas. By altering the way noise is added to images, we can shift from adding noise to the entire image to adding noise to specific key objects. 
It can enable the model to specifically improve its focus on image information in certain domains, thus having extensive industrial applications. 
Besides, we believe that the core part of the TPO method, the visual-anchored reward scoring method, possesses strong extensibility. For example, these token-level rewards can also be used to weight the probability distributions in the calculations for other RLHF methods, enhancing the visual attention of multimodal models.

We will continue to expand in this direction, and we believe that the technology we have proposed in this paper has a vast space for further development and application.

\section{Ethic Statement}
The main purpose of this article is to alleviate the hallucination problem in LVLM using reinforcement learning method. By employing a self-calibrated visual-anchored reward approach, we propose the TPO method, which significantly addresses the hallucination issue and helps the model connect with more visual information. 
All the models and datasets we used are open source, so we believe that the work in this paper does not pose any potential threats.

\bibliography{anthology,custom}

\appendix

\begin{table*}[t]
    \centering
    \fontsize{9.5pt}{12.5pt}\selectfont
    \begin{tabular}{p{2.3cm}p{0.7cm}p{0.7cm}p{0.8cm}p{0.85cm}p{0.75cm}p{0.75cm}p{0.8cm}p{0.5cm}p{0.6cm}p{0.7cm}p{1.2cm}}  
        \hline
        \multirow{2}{*}{\textbf{Method}} & \multicolumn{2}{c}{\textbf{AMBER}} & \multicolumn{2}{c}{\textbf{MMHal}} & \multicolumn{3}{c}{\textbf{HallusionBench}} &\multicolumn{4}{c}{\textbf{General Benchmarks}}\\
        \cmidrule(r){2-3}  \cmidrule(r){4-5} \cmidrule(r){6-8} \cmidrule(r){9-12}
        & Acc  & F1  & Score  & Hal $\downarrow$
        & Easy& Hard& aAcc  & SEED & MMB & LLaVA & MM-Vet\\
        \hline
        LLAVA-1.5-7B & 71.7	&74.3	&2.01	&61.46&	42.64	&41.16	&47.21 & 66.1	 &\underline{73.3}	 &65.6	 &31.6 \\
        \hline
        0 setp& 77.6&	82.6&	2.10&	58.33&	\textbf{44.40}&	45.35&	49.42&66.2	&73.2&	69.9&	32.1\\
        250 steps & 79.0&	84.5&	2.33	&53.13&	43.52	&46.05&	49.51& 66.6	&73.4&	68.5&	31.3\\
        750 steps &79.30 &	85.0	&2.40&	52.08&	41.76	&48.14&	50.04 & \textbf{66.7}& 	73.5& 	69.2& 	32.8\\
        999 steps &79.20	&85.0	&2.41&	52.08&	41.76&	47.67&	49.69&\textbf{66.7}	&73.5&	69.2&	\textbf{33.3}\\
        \hline
         \rowcolor[gray]{0.9}\textbf{500 steps (Ours)}& \textbf{79.30}&	\textbf{85.0}&	\textbf{2.47}&	\textbf{51.04}&	41.76&	\textbf{48.37}&	\textbf{50.22}&	66.6&	\textbf{73.6}&	\textbf{70.2}&	33.0\\
        \hline
    \end{tabular}
    \caption{Detail of Figure \ref{fig:step} (a).}
    \label{table.4}
\end{table*}

\begin{table*}[t]
    \centering
    \fontsize{9.5pt}{12.5pt}\selectfont
    \begin{tabular}{p{2.3cm}p{0.7cm}p{0.7cm}p{0.8cm}p{0.85cm}p{0.75cm}p{0.75cm}p{0.8cm}p{0.5cm}p{0.6cm}p{0.7cm}p{1.2cm}}  
        \hline
        \multirow{2}{*}{\textbf{Method}} & \multicolumn{2}{c}{\textbf{AMBER}} & \multicolumn{2}{c}{\textbf{MMHal}} & \multicolumn{3}{c}{\textbf{HallusionBench}} &\multicolumn{4}{c}{\textbf{General Benchmarks}}\\
        \cmidrule(r){2-3}  \cmidrule(r){4-5} \cmidrule(r){6-8} \cmidrule(r){9-12}
        & Acc  & F1  & Score  & Hal $\downarrow$
        & Easy& Hard& aAcc  & SEED & MMB & LLaVA & MM-Vet\\
        \hline
        LLAVA-1.5-7B & 71.7	&74.3	&2.01	&61.46&	42.64	&41.16	&47.21 & 66.1	 &\underline{73.3}	 &65.6	 &31.6 \\
        \hline
        $a=0$ &	79.2&	83.0&	2.24&	56.25&	\textbf{42.20}&	43.72&	48.27& 66.6	&73.5	&68.4&	32.8\\
        $a=1$ &79.2&	84.9&	2.44&	\textbf{48.96}&	41.54&	47.44&	49.60 &\textbf{66.7}	&\textbf{73.6}&	\textbf{70.8}	&\textbf{33.1}\\
        \hline
         \rowcolor[gray]{0.9}\textbf{$a=0.5$ (Ours)}& \textbf{79.3}&	\textbf{85.0}&	\textbf{2.47}&	51.04&	41.76&	\textbf{48.37}&	\textbf{50.22}&	66.6&	\textbf{73.6}&	70.2&	33.0\\
        \hline
    \end{tabular}
    \caption{Detail of Figure \ref{fig:step} (b).}
    \label{table.5}
\end{table*}

\section{Implement Details}\label{A}
\subsection{Setup}
\label{app_set}
In our experiments, we trained the LLaVA-v1.5 model. For our TPO method and the vanilla DPO method, we set the maximum learning rate to 5e-8 on the 7B version and trained for 4 epochs. We set the maximum learning rate to 2e-7 on the 13B version and trained for 4 epochs. The RLHF-V training was set according to the paper \cite{yu2024rlhf}. All parts requiring GPT-4 evaluation use the GPT-4-0613 8K version, and the MM-Vet testing is conducted on the official evaluation website. 

During the training process, we froze the vision encoder and only trained the LLM.

We also trained the Qwen2-VL model. For our TPO method and the vanilla DPO method, we set the maximum learning rate to 5e-9 for 7B model, 1e-9 for 2B model and trained for 4 epochs.

For a fair comparison, we set the seed to 42 during training and greedy decoding was used during inference.

Our experiments were all conducted on a server equipped with 8 Nivdia A100 GPUs; in specific cases (such as the 13B model), we utilized 32 Nivdia A100 GPUs. For the hyperparameter settings, all hyperparameters are consistent with those of our main experiment. 
We used `Pytorch' in our code.
Moreover, the level of diffusion noise in our model is represented by a formula $\xi= Sigmoid(l_t) \times (0.5 \times 10^{-2} - 10^{-5}) + 10^{-5}$, where $l_t$ is a list of 1,000 numbers taken at equal intervals over the interval $[-6, 6]$, and $\epsilon \in N(0,1)$.

The cases in Figure \ref{fig:teaser} and Figure \ref{fig:attention} come from benchmarks \cite{sun2023aligning}, while the cases in Figure \ref{fig.2} come from the RLHF-V training set \cite{yu2024rlhf}.
\subsection{Benchmarks}
The three hallucination benchmarks: (1) AMBER : a multi-dimensional hallucination benchmark with more than 15K samples, including discriminative and description tasks. (2) MMHal-Bench : it measures the hallucination rate and informativeness of responses. (3) HallusionBench : it evaluates visual illusions and knowledge hallucinations through systematically structured discriminative tasks. 

The four general benchmarks: (1) SEED Bench : a benchmark for LVLMs on generative comprehension. (2) MMBench: a comprehensive benchmark designed to evaluate the capabilities across various tasks and modalities. (3) LLaVA Bench: a benchmark for evaluating multi-modal conversation, detailed description, and complex reasoning. (4) MM-Vet: a benchmark to assess integrated capabilities.

\subsection{Training Efficiency}
In TPO, generating corrupted images at each step incurs almost no time cost, as it is done during the initial data preparation. The main time consumption comes from calculating logits $p_{log}(y_i | x,v_{c},y_{<i})$ for the noisy images. 

We have also conducted a careful analysis of the time consumption for LLava-1.5-7B under the settings in Section~\ref{app_set}, the training durations for DPO and TPO were 1 hour 24 minutes and 1 hour 57 minutes, respectively, indicating about a 40\% increase in time. Nevertheless, all training methods aimed at eliminating hallucinations inevitably incur additional time costs, compared to other methods requiring fine-grained annotations, our self-calibrated approach with 40\% time increase proves to be sufficiently efficient. 

It has also shown superior outcomes on 5K training data training to CSR training on 13K data and POVID training on 17K data. This highlights the efficacy of our method in guiding the model to pay more attention to image details and in reducing hallucinations.

\begin{lstlisting}[caption={Example Python Code for Noise Addition}, numbers=none, label={lst:noise-addition}]
pil_to_tensor = transforms.ToTensor()
tensor_to_pil = transforms.ToPILImage()
image = Image.open(default_image_path).convert("RGB")
image_tensor = pil_to_tensor(image)
image_noisy = add_diffusion_noise(image_tensor, 500)
image_noisy = tensor_to_pil(image_noisy)
\end{lstlisting}

\section{Comparison with Related Methods}
To more comprehensively highlight the advantages of the TPO method, we conducted comparisons with other related works \cite{jiang2024hallucination,yue2024less,xiao2025detecting, sarkar2024mitigating, zhao2023beyond,leng2024mitigating,huang2024opera,zhou2023analyzing} aimed at addressing the hallucination problem.
The results show that TPO achieves more significant hallucination reduction.


Preference alignment and decoding strategies are two important and parallel categories of methods for hallucination mitigation. We believe that training with preference alignment offers several advantages: 1) Direct Optimization of Output Preferences: This approach directly optimizes the model's output to align with desired preferences without requiring changes to the decoding strategy. 2) Higher Inference Efficiency: Preference alignment typically results in more efficient inference, as it does not introduce additional complexity during the decoding process.
 
One key advantage of decoding methods is that they do not require retraining the model, making them highly efficient for deployment. However, this does not preclude the benefits of preference alignment. In fact, we believe combining these two approaches can yield even better results.

To further elucidate the innovations of our work, we provide a more in-depth comparison with similar methods \cite{cui2024fine,zhou2024aligning,leng2024mitigating}. Our approach exhibits both conceptual distinctions and methodological advancements. Unlike FiSAO, which relies on a vision encoder to compute token-level scores, TPO derives them directly from the LVLM’s logits difference between clean and noisy images. Moreover, while FiSAO builds upon PPO, TPO enhances the DPO framework. Our method shares with VCD the idea of using noise-induced logits change to identify visual-sensitive tokens; however, TPO goes beyond identification by introducing a self-calibrating reward Eq.~\ref{(6)} that jointly reinforces positive outputs, suppresses hallucinations, and strengthens image–token alignment—aiming fundamentally to improve perceptual faithfulness. In contrast, VCD focuses on decoding less image-affected tokens without enhancing underlying visual grounding. Similarly, although POVID also uses Gaussian noise for perturbation, it primarily constructs negative samples for DPO training, whereas TPO leverages noise to quantify token–image relevance and enables a more efficient token-level reinforcement learning strategy, leading to superior performance.

\section{Results on Object Hallucination}\label{E}
In the AMBER benchmark, there is a subset for evaluating object hallucinations in image description tasks. Since this paper focuses on visual question answering, this part of the experiment is included in this section. To assess the proportion of object hallucinations in image descriptions, AMBER uses \textbf{Chair} as the metric. 

\begin{figure}[t]
\begin{center}
\includegraphics[scale=0.5]{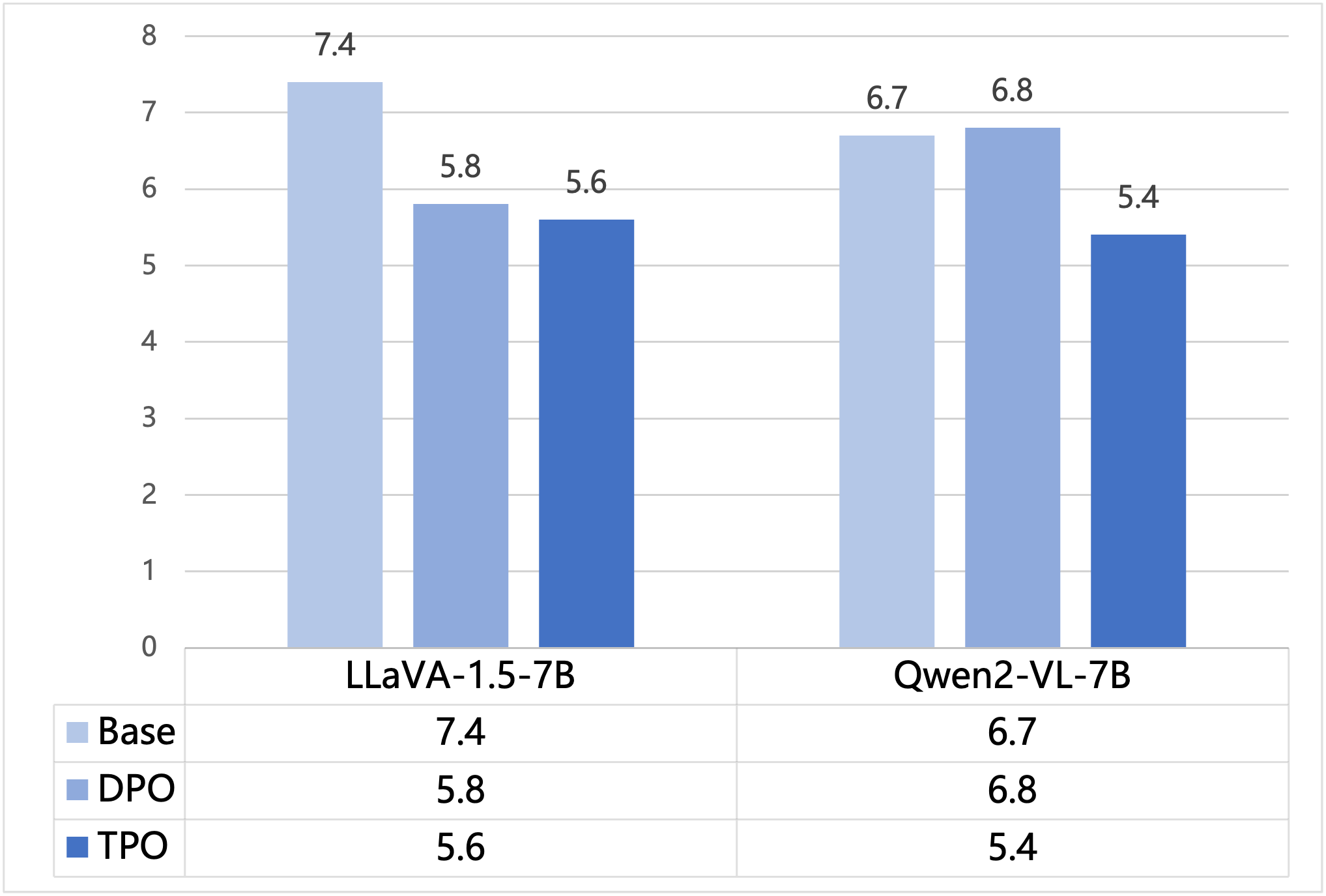}
\caption{Chair Performance Comparison.}
\label{chair}
\end{center}
\end{figure}

The results are shown in Figure~\ref{chair}. Note that 'Chair' represents the hallucination ratio, where a smaller value indicates better model performance. To more clearly illustrate the comparison between methods in the figure, we use $10-chair$ as the indicator. The results show that TPO can not only mitigate the hallucination in visual question answering, but also eliminate the object hallucination in image descriptions to a certain extent.
\begin{table}[t]
    \centering
   \small
    \begin{tabular}{lcccc}
        \hline
     
        \multirow{2}{*}{\textbf{Method}}& \multicolumn{2}{c}{\bf AMBER} & \multicolumn{2}{c}{\bf MMHal} \\
        \cmidrule(r){2-3}  \cmidrule(r){4-5}
        & Acc  & F1   & Score & Hal$\downarrow$ \\
        \hline 
        LLaVA-1.5-7B & 71.7 & 74.3 & 2.01 & 61.46 \\
        VCD        & 71.8 & 74.9 & 2.12 & 54.20 \\
        LURE       & 73.5 & 77.7 & 1.64 & 60.40 \\
        OPERA      & 75.2 & 78.3 & 2.15 & 54.20 \\
        HACL	&2.13	&50&	-&	-\\
        EOS&	2.03&	59	&-&	-\\
        HA-DPO	&1.97	&60&	75.2&	79.9\\
        HALVA	&2.25&	54&	-	&83.4\\
        DPO       & 77.5 & 82.1 & 2.14 & 58.33 \\
        TPO       & \textbf{79.3} & \textbf{85} & \textbf{2.47} & \textbf{51.04} \\
        \hline 
        LLAVA-1.5-13B	&2.38&	53	&71.3	&73.1\\
HSA-DPO	&2.61&	48&	-&	-\\
HALVA&	2.58&	\textbf{45}&	-&	86.5\\
DPO&	2.47&	51	&83.2	&86.9\\
TPO&	\textbf{2.72}	&46&	\textbf{83.9}&	\textbf{88}\\
        \hline
    \end{tabular}
    \caption{Comparison of Results}
    \label{tab:comparison}
\end{table}

\begin{table}[t]
    \centering
    \fontsize{9pt}{12pt}\selectfont
    \begin{tabular}{p{1.34cm}p{0.42cm}p{0.42cm}p{0.42cm}p{0.42cm}p{0.42cm}p{0.42cm}p{0.42cm}}
        \hline
        \multirow{2}{*}{\textbf{Method}} & \multicolumn{2}{c}{\textbf{AMBER}} & \multicolumn{2}{c}{\textbf{MMHal}} & \multicolumn{3}{c}{\bf HallusionBench} \\
        \cmidrule(r){2-3}  \cmidrule(r){4-5} \cmidrule(r){6-8}
                      & Acc  & F1  & Score & Hal$\downarrow$ & Easy  & Hard & aAcc \\
        \hline
        LLaVA-1.5-7B      & 71.7& 74.3& 2.01& 61.5& 42.6 &41.2 &47.2 \\
        \hline
        +TPO (white)&78.0&	82.7&	2.26&	55.2&	\textbf{44.2}&	45.4&	49.3 \\
        \rowcolor[gray]{0.9} \textbf{+TPO}& \textbf{79.3}& \textbf{85.0}& \textbf{2.5}& \textbf{51.0}& 41.8& \textbf{48.4}& \textbf{50.2} \\
        \hline
    \end{tabular}
    \caption{Comparison of different noise adding method. “white" indicates that blank images are used in place of noisy images.}
    \label{white}
    \vspace{-0.5cm}
\end{table}

\section{Comparison of Different Noise Adding Methods.}\label{F}
To evaluate the impact of different methods of adding noise to images on our approach, we test a scheme where noise images were replaced with white images under the same experimental conditions. The results, shown in Table \ref{white}, demonstrate the superior performance of our method. We believe that the noise addition method used in our paper can control noise levels to create images that are more likely to induce hallucinations in the model, thereby achieving better results.

\section{Evaluation of TPO Training on Real-World Performance}
To validate the effect of TPO training, we compared the performance of the LLaVA-1.5-7B models before and after training on a real-world image set of 10 images, as suggested during the review process. A manual analysis of the outputs revealed that the TPO-trained model provided superior results in 40\% of the cases. In contrast, the original model was better in only 20\% of cases, with both models performing equally well in the remaining 40\%. This represents a clear net performance gain attributable to the TPO training procedure.

\begin{table}[t]
    \centering
    \fontsize{9.5pt}{12.5pt}\selectfont
    \begin{tabular}{p{2.3cm}cc} 
        \hline
        \multirow{2}{*}{\textbf{Method}} & \multicolumn{2}{c}{\textbf{AMBER}} \\
        \cmidrule(r){2-3} 
        & Acc  & F1 \\
        \hline
        LLaVA-1.5-7B & 71.7 & 74.3 \\
        + DPO & 77.5 & 82.1 \\
        + CSR & 73.2 & 76.1 \\
        + POVID & 71.9 & 74.7 \\
        + RLHF-V & 74.8 & 78.5 \\
        \rowcolor[gray]{0.9} + \textbf{TPO (1K)} & 72.5 & 75.3 \\
        \rowcolor[gray]{0.8} + \textbf{TPO (3K)} & 78.9 & 83.4 \\
        \rowcolor[gray]{0.7} + \textbf{TPO (5K)} & \textbf{79.3} & \textbf{85.0} \\
        \hline
    \end{tabular}
    \caption{Evaluation of TPO Training with varying amounts of training data.} 
    \label{table:9} 
\end{table}

\section{Analysis of TPO Training on less training data}
In order to further investigate the impact of training data volume, we constructed subsets of 1K and 3K samples from the original 5K training set. Using these subsets, we fine-tuned the LLaVA-1.5-7B model with a consistent learning rate of 5e-8 for 4 epochs, maintaining identical experimental settings as in the full-data scenario. 

As anticipated in table \ref{table:9}, the performance of the TPO method exhibits a gradual decline with reduced training data. Nevertheless, even when trained on only 1K samples, TPO remains competitive and surpasses most baseline methods. With 3K training samples, it achieves the second-highest performance, further underscoring the effectiveness and data efficiency of our approach.

\end{document}